\documentclass[letterpaper, 10 pt, conference]{ieeeconf}  

\IEEEoverridecommandlockouts                              

\usepackage{graphics} 
\usepackage{epsfig} 
\usepackage{amsmath} 
\usepackage{amssymb}  
\usepackage{cite}
\usepackage{color}
\usepackage{balance}
\usepackage{booktabs}

\usepackage{algorithm}
\usepackage{algorithmic}

\usepackage{multirow}
\usepackage{adjustbox}
\usepackage{url}

\makeatletter
\let\MYcaption\@makecaption
\makeatother

\usepackage[font=footnotesize]{subcaption}

\makeatletter
\let\@makecaption\MYcaption
\makeatother

\usepackage{subcaption}

\DeclareMathOperator*{\argmax}{arg\,max}

\title{\LARGE \bf
BenchNav: Simulation Platform for Benchmarking Off-road Navigation Algorithms with Probabilistic Traversability
}

\author{Masafumi Endo$^{1}$, Kohei Honda$^{2}$, and Genya Ishigami$^{1}$
\thanks{*This work was partially supported by JSPS KAKENHI Grant Number JP22J22731.}
\thanks{$^{1}$M. Endo and G. Ishigami are with the Space Robotics Group, Department of Mechanical Engineering, Keio University, Kanagawa 223-8522, Japan
        {\tt\small 
        masafumi.endo@keio.jp,
        ishigami@mech.keio.ac.jp}
        }
\thanks{$^{2}$K. Honda is with the Mobility System Group, Department of Mechanical Systems Engineering, Nagoya University, Aichi 464-8603, Japan 
        {
        \tt\small
        honda.kohei.f4@a.mail.nagoya-u.ac.jp
        }
        }%
}

\begin{document}

\maketitle

\thispagestyle{empty}
\pagestyle{empty}

\begin{figure*}[t]
    \centering
    \includegraphics[width=0.85\hsize]{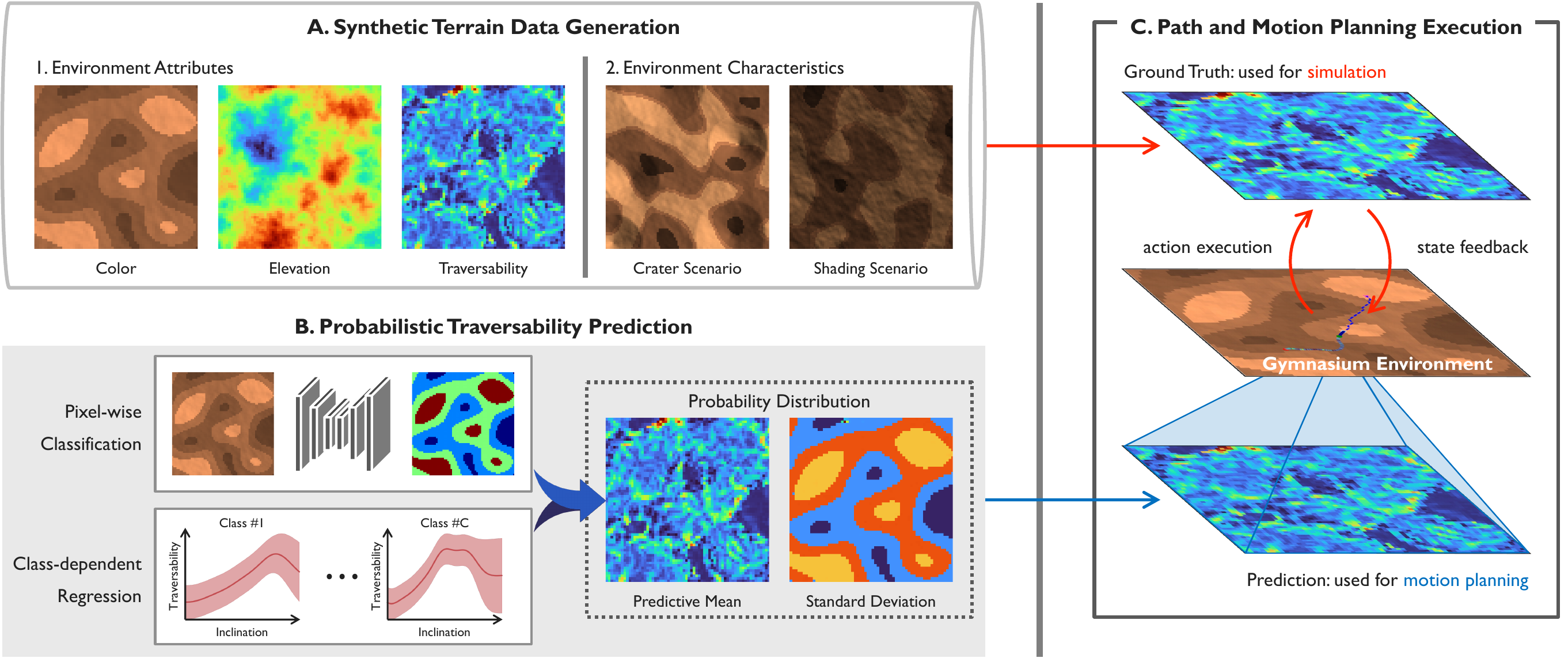}
    \caption{BenchNav, a platform designed to solve off-road navigation problems, consists of A) synthetic terrain data generation, B) probabilistic traversability prediction based on the classify-then-regress method, and C) path and motion planning execution with given problem formulations and solvers.}
    \vspace{-5pt}
    \label{fig:1}
\end{figure*}

\begin{abstract}
As robotic navigation techniques in perception and planning advance, mobile robots increasingly venture into off-road environments involving complex traversability.
However, selecting suitable planning methods remains a challenge due to their algorithmic diversity, as each offers unique benefits.
To aid in algorithm design, we introduce BenchNav, an open-source PyTorch-based simulation platform for benchmarking off-road navigation with uncertain traversability.
Built upon Gymnasium, BenchNav provides three key features: 1) a data generation pipeline for preparing synthetic natural environments, 2) built-in machine learning models for traversability prediction, and 3) consistent execution of path and motion planning across different algorithms. 
We show BenchNav's versatility through simulation examples in off-road environments, employing three representative planning algorithms from different domains.\\
\url{https://github.com/masafumiendo/benchnav}
\end{abstract}

\section{Introduction}

Mobile robots unchain us from labor-intensive, hazardous activities \emph{in the field}, ranging from environmental monitoring~\cite{matthew2012robots}, search and rescue~\cite{nagatani2013emergency}, to planetary exploration~\cite{vandi2023autonomous}.
Autonomous navigation is critical for reliable decision-making, enabling safe and efficient mission operations without consistent human intervention.
However, off-road navigation is fraught with risks in unstructured terrains, such as robots getting stuck in deformable terrain, tipping over on steep slopes, or colliding with both positive and negative obstacles.
\emph{Traversability} is thus an essential criterion to quantify the level of robot ability to traverse, ranging from safe to hazardous, in unstructured off-road environments.

Off-road navigation has evolved with path and motion planning algorithms that incorporate traversability either through theoretical or machine learning (ML)-based modeling.
\emph{Search-based} path planning, valid in discretized graph environments, is notable for its resolution optimality.
It can generate global paths for navigation in extreme environments, such as planetary surfaces~\cite{goldberg2002stereo,ishigami2007path,daftry2022mlnav,endo2023risk,vandi2023autonomous} and subterranean domains~\cite{fan2021step}, making it suitable for real-world applications, including NASA's AutoNav~\cite{goldberg2002stereo} and Enav~\cite{vandi2023autonomous} systems.
Nevertheless, search-based methods often miss robot dynamics, leading to challenges in translating global paths to local motions.
\emph{Sampling-based} motion planning has become another common approach, capable of handling high-dimensional spaces through random environmental sampling to construct feasible robot states.
Its flexible representation in configuration space allows us to apply complex dynamical models in motion planning, such as closed-loop rapidly-exploring random trees (CL-RRT)~\cite{kuwata2009real} and its variants in off-road navigation~\cite{takemura2021traversability,takemura2023uncertainty}.
Still, they often indirectly include dynamical models and require substantial computational effort to reach asymptotic optimality.
\emph{Optimization-based} algorithms, in contrast, enable the direct integration of dynamical models into motion planning, resulting in optimal state and control sequences.
Model predictive path integral control (MPPI)~\cite{williams2017information}, a sampling-based optimal control solver, is favored for aggressive off-road navigation~\cite{gasparino2022wayfast,gasparino2024wayfaster,gibson2023multi,han2023model,cai2022risk,cai2022probabilistic,cai2023evora}, offering rapid trajectory optimization and the ability to handle non-differentiable objectives.
These techniques focus on generating sequential, short-horizon solutions while often failing to achieve global optimality.

We here argue that off-road autonomy requires a well-aligned integration of planning algorithms and traversability modeling for robot dynamics. 
Along with the advances in ML models and growing dataset availability~\cite{triest2022tartandrive,sharma2022cat,sivaprakasam2023tartandrive,sivaprakasam2024tartandrive}, many studies push forward data-driven traversability prediction and its application to motion planning.
However, the lack of standardized benchmarking for planning raises the question: \emph{How can we select the suitable algorithm for designing an off-road navigation system, given their unique differences?}
This is often empirically tailored to specific scenarios, though quantitative comparison across different domains is a tedious yet necessary process.

To address the above challenge, we present BenchNav, an open-source, PyTorch-based~\cite{paszke2019pytorch} simulation platform, designed for \textbf{Bench}marking off-road \textbf{Nav}igation algorithms.
Leveraging Gymnasium~\cite{brockman2016openai}, BenchNav equips three key features to tackle off-road navigation problems as follows:
\begin{itemize}
    \item Synthetic data generation replicates natural terrain, embedding the challenges of traversability prediction,
    \item Built-in ML models for probabilistic traversability prediction from given environmental data,
    \item Path and motion planning algorithm executions unified with robot models factoring in uncertain traversability.
\end{itemize}
The synthesized process, encompassing data preparation, traversability modeling, and planning, ensures easy and consistent navigation simulations across various algorithms.
We perform simulation experiments in deformable off-road environments, focusing on vehicle slip as a typical metric for traversability modeling.
The use of different path and motion planning algorithms demonstrates BenchNav's versatility.

\section{Problem Statement}

\label{section:problem_statement}

\emph{We state the off-road navigation problem for mobile robots as one embracing traversability prediction and motion planning.}
Unstructured environments present complex physical interactions with robots, leading to deviations between commanded and actual velocities.
Under significant disturbances, the discrete-time system is given as $\mathbf{s}_{t+1}=F\left(\mathbf{s}_t, \mathbf{a}_t,\boldsymbol{\lambda}\right)$, where $t$ denotes discrete time steps.
Here, $\mathbf{s}_t \in \mathbb{R}^n$ represents the state vector, $\mathbf{a}_t \in \mathbb{R}^m$ the action vector serving as the control input to the robot, and $\boldsymbol{\lambda} \in [0, 1]^p$ the traversability coefficient vector, scales $\mathbf{a}_t$.
The state transition function $F$ integrates these inputs to update the state vector, determining the system's response accounting for terrain interactions.

The traversability prediction objective is to create a model $\mathcal{M}$, mapping environmental observations $\mathbf{o} \in {O}$ to traversability coefficients $\boldsymbol{\lambda} \in {\Lambda}$, where ${O}$ and ${\Lambda}$ are sets of all environmental observations and traversability coefficients, respectively.
ML is a powerful tool for discovering latent models by learning general correlations from training data.
Recent studies also reveal the value of quantifying uncertainty in traversability learning~\cite{endo2022active,fan2022learning,triest2023learning}, arising from inherent observation noise and limited training data.
We thus adopt probabilistic ML models as the basis for traversability prediction used across planning algorithms.

Besides, the motion planning objective is to find a sequence of feasible actions $\mathbf{a}_{t} = \pi (\mathbf{s}_t, \boldsymbol{\hat{\lambda}})$ towards its destination, with a policy $\pi$ that translates current states and traversability coefficients into the state transition dynamics.
The motion planning problem is formulated to optimize user-specified objectives, such as time efficiency, subject to certain constraints.
Despite the various solvers available, we emphasize the need to exploit uncertainty in traversability prediction.
Following earlier studies~\cite{cai2022risk, cai2022probabilistic, cai2023evora, endo2023risk}, we employ statistical methods for risk inference, such as conditional value at risk (CVaR)~\cite{majumdar2020how}, to transform probability distributions into deterministic inputs.
This approach bridges traversability prediction and motion planning, ensuring consistent execution in solving the off-road navigation problem.
\section{BenchNav}

This section details BenchNav, a simulation platform tailored for off-road navigation.
We developed BenchNav on top of Gymnasium (formerly OpenAI Gym), leveraging its capabilities for sequential decision-making and scalable simulation of diverse environments.
BenchNav's implementation in PyTorch further enables GPU acceleration, ensuring efficient computation for ML models and planning algorithms throughout the platform.
The overall simulator pipeline, depicted in Fig. \ref{fig:1}, unfolds as follows:
It first generates controlled datasets containing ${O}$ and their respective ${\Lambda}$, engineered to catch the difficulties present in traversability prediction.
Then, a set of all probability distributions for traversability coefficients, denoted as $\mathbb{P}({\hat{\boldsymbol{\lambda}}}|\mathbf{o})$, is formed by initially categorizing symbolic terrain classes from appearance features, followed by predicting class-dependent latent traversability (LT) functions from geometric features.
The final stage involves running planning algorithms to find $\mathbf{a}_{t} = \pi (\mathbf{s}_t, \hat{\boldsymbol{\lambda}})$, with given problem formulations and solvers that incorporate risk inference metrics for uncertain traversability.

\subsection{Synthetic Terrain Data Generation}

We synthesize a top-down 2.5D terrain map instance, replicating pixel-wise appearance and geometric features, alongside their latent traversability coefficients.
In off-road scenarios, appearance features indicate surface properties, determining distinct traversability trends among terrain classes.
Geometric features also dominate the variation in traversability within each class.
The data is thereby paired: RGB color data with their corresponding terrain classes $c \in C$, where $C$ is the set of all terrain classes, and elevation data coupled with class-dependent LT functions $f_{c}(\psi)$, where $\psi$ is the inclination derived by the Horn method~\cite{ono2018mars}.

For each map instance, occupancy ratios dictate terrain class distribution; we create such a distribution with Perlin noise to assign color and terrain classes based on the clusters.
We then simulate rough terrain elevation with fractal terrain modeling~\cite{yokokohji2004evaluation} to compute $\lambda = f_{c}(\psi) + \epsilon_{c}$, where $\epsilon$ represents additive zero-mean Gaussian noise, and to apply shading to the color data due to elevation changes.
We consequently get a map of colors, elevations, and traversability coefficients, depicted in Fig.~\ref{fig:1}A1.
Environmental features are readily adjusted to simulate challenges in traversability prediction, such as ambiguous appearance from intense shading and irregular geometry in crater-like terrains, shown in Fig.~\ref{fig:1}A2.

\subsection{Probabilistic Traversability Prediction}

BenchNav employs two types of pre-trained ML models: 1) a terrain classifier for predicting pixel-wise terrain classes and 2) Gaussian processes (GPs)~\cite{williams2006gaussian} for estimating class-dependent LT functions with uncertainties.
These distinctive ML models are integrated into a single probability distribution via mixtures of GPs~\cite{tresp2000mixture}, representing probabilistic traversability at $\mathbf{s}$ as follows:
\begin{equation}
    \label{eq:1}
    \mathbb{P}_{\mathbf{s}}(\hat{\lambda})=\sum_{c \in C} \mathbb{P}_{\mathbf{s}}\left(c\right) \mathbb{P}_{c}(\hat{\lambda} \mid \psi_{\mathbf{s}}),
\end{equation}
where $\mathbb{P}_{c}(\hat{\lambda} \mid \psi_{\mathbf{s}})$ denotes class-dependent GPs, weighted by a categorical distribution $\mathbb{P}_{\mathbf{s}}(c)$ from the classifier.
This probabilistic fusion~\cite{endo2023risk} constructs $\mathbb{P}({\hat{\boldsymbol{\lambda}}}|\mathbf{o})$, as shown in Fig.~\ref{fig:1}B.

While any model capable of predicting pixel-wise categorical distributions is suitable for terrain classification, we opt for the U-Net model~\cite{ronneberger2015unet} with a ResNet-18 backbone~\cite{he2016deep}.
Users also have the option of selecting a single GP that corresponds to the most likely class $c^* = \argmax_c\mathbb{P}_{\mathbf{s}}(c)$ in eq.~(\ref{eq:1}), rather than summing over multiple GPs.

\subsection{Path and Motion Planning Execution}

The last stage simulates off-road robot navigation from a start state $\mathbf{s}_{\text{start}}$ to a goal state $\mathbf{s}_{\text{goal}}$ within the allotted finite positive time budget $T$. The motion planning problem in off-road environments is given as follows:
\begin{subequations}
\begin{align}
    \text{Find: } & \left\{\mathbf{a}_t\right\}_{t=0}^{T-1}, \; 
    \left\{\mathbf{s}_t\right\}_{t=0}^{T} \\
    \text{Minimize: } & J\left(\left\{\mathbf{a}_t\right\}_{t=0}^{T-1}, \;
    \left\{\mathbf{s}_t\right\}_{t=0}^T \right) \\
    \text{subject to: } & \mathbf{s}_0 = \mathbf{s}_\text{start}, \; \mathbf{s}_T = \mathbf{s}_\text{goal} \\
                        & \mathbf{s}_t \in S_\text{free}, \; ^\forall t \in \left\{0, \ldots, T\right\} \\
                        & \mathbf{s}_{t+1} = F(\mathbf{s}_t, \mathbf{a}_t, \hat{\boldsymbol{\lambda}}), \; ^\forall t \in \left\{0, \ldots, T-1\right\}
    \label{eq:2e}
\end{align}
\end{subequations}
$J$ is the cost function quantifying trajectory optimality to be minimized.
$S_{\text{free}}$ denotes the set of all free states, indicating navigable space free from hazards.
Solving the above problem yields the optimal action vector $\mathbf{a}^*_t$ at each $t$, controlling the robot to its destination. 
We design BenchNav to solve the given optimization problem in an iterative fashion, allowing for real-time feedback that compensates for modeling errors in $F$.
Users can define $F$, $J$, and $S_{\text{free}}$ to formulate the problem according to their respective objectives.
BenchNav simulates navigation with deployed planning algorithms as solvers, including both global and local combinations.
Here, risk inference metrics convert $\mathbb{P}(\hat{\boldsymbol{\lambda}}|\mathbf{o})$ to $\hat{\boldsymbol{\lambda}}$, incorporating uncertainty in traversability prediction.
BenchNav continuously monitors the robot's state, including its experience with traversability, to observe how it safely reaches its destination within $T$, which is then marked as successful navigation.

\begin{figure*}[t]
	\begin{minipage}[t]{0.33\linewidth}
		\centering
		\includegraphics[clip, width=47.5mm]{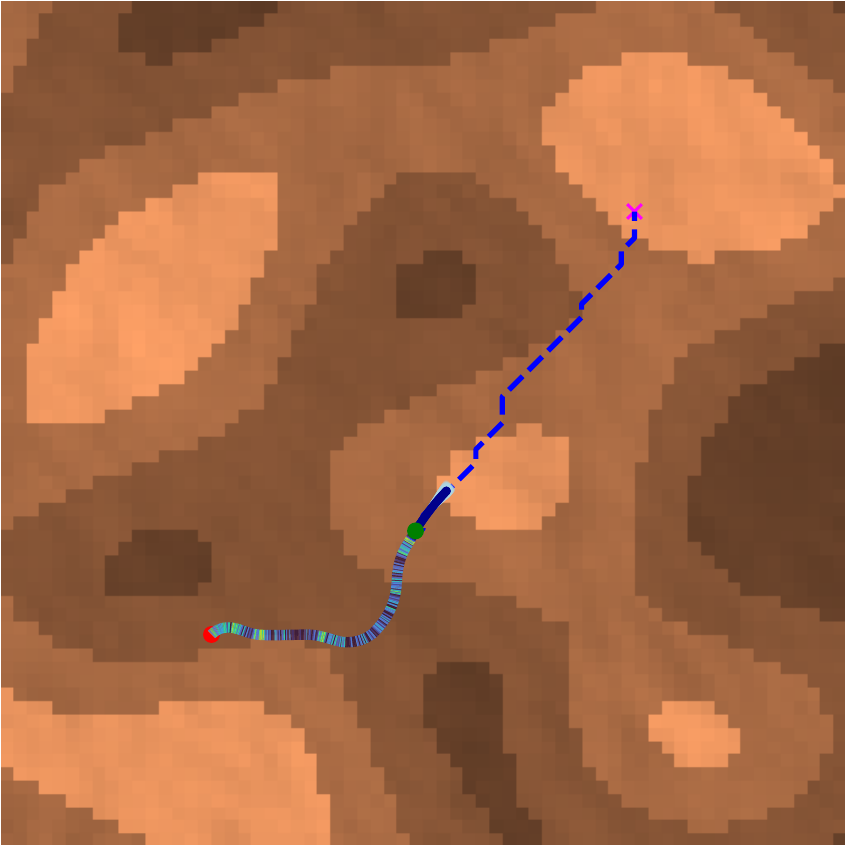}\\
		\subcaption{\textit{Search-based}: A*+DWA}
	\end{minipage}
	\begin{minipage}[t]{0.33\linewidth}
		\centering
		\includegraphics[clip, width=47.5mm]{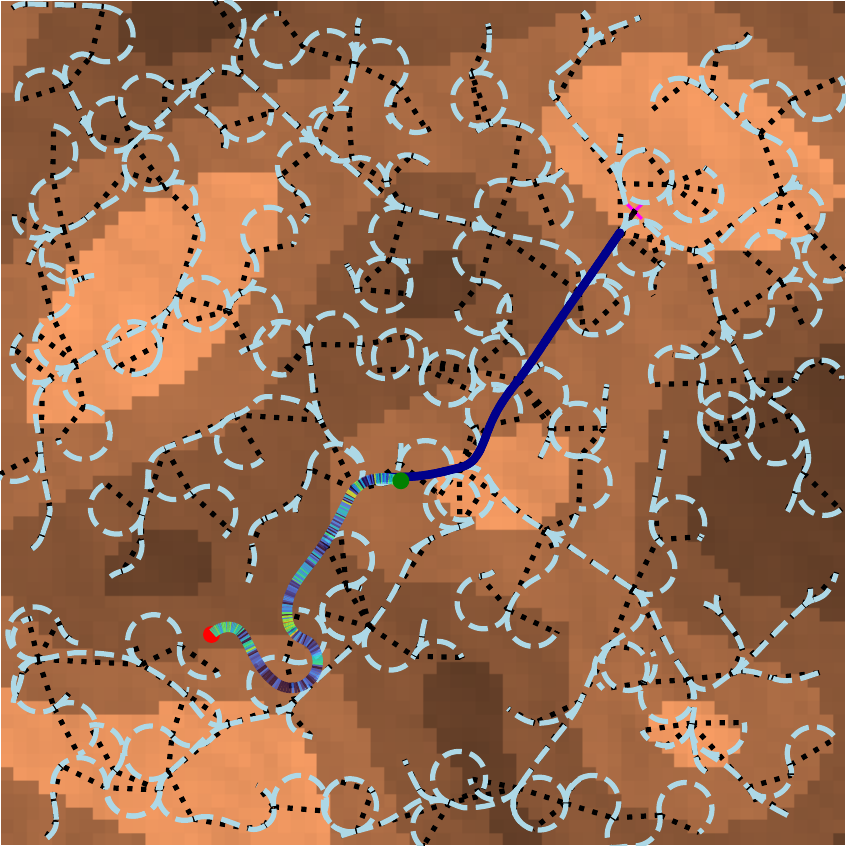}\\
		\subcaption{\textit{Sampling-based}: CL-RRT}
	\end{minipage}
	\begin{minipage}[t]{0.33\linewidth}
		\centering
		\includegraphics[clip, width=47.5mm]{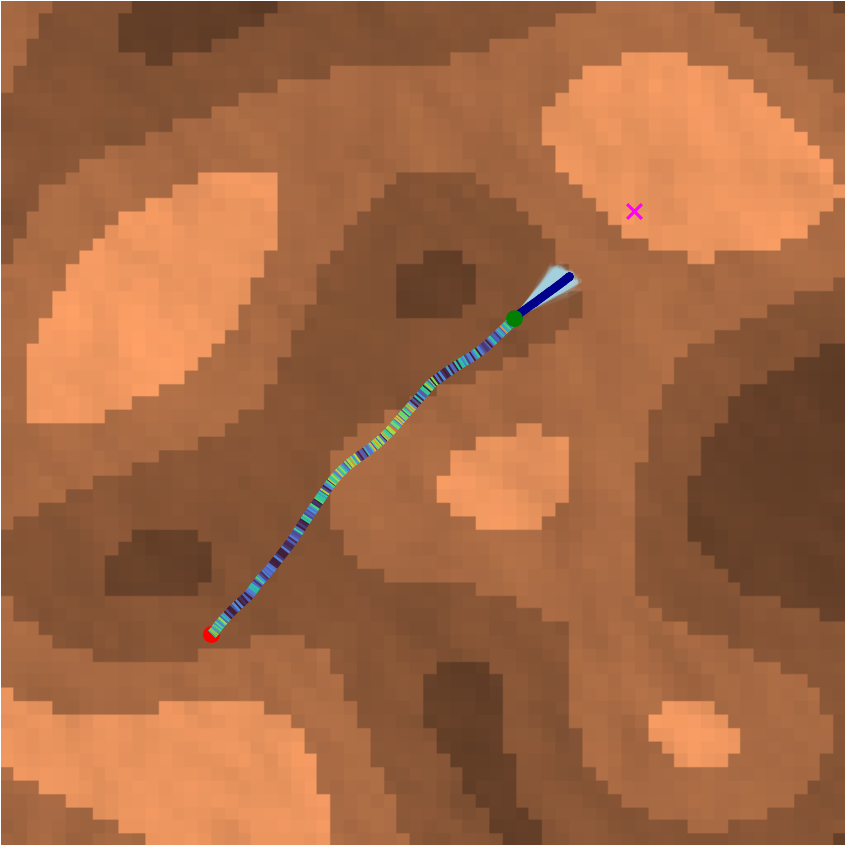}\\
		\subcaption{\textit{Optimization-based}: MPPI}
	\end{minipage}
    \caption{Off-road navigation snapshots with different planning algorithms at $t=25\, [\text{sec}]$. Red circles, magenta crosses, and green circles mark the start, goal, and current robot positions, respectively. Trajectories are color-coded according to traversability, with cooler colors for safer paths. (a) A* reference paths are blue dashed lines; DWA results are shown with a dark blue solid line and light blue lines. (b) Black dashed lines show edges; light blue dashed lines are for closed-loop simulations, with solid blue lines for planned trajectories. (c) MPPI outcomes are a dark blue solid line with light blue top samples.}
    \vspace{-5pt}
\label{fig:2}
\end{figure*}

\section{Simulation Experiments}

\subsection{Implementation Details}

We test BenchNav's versatility through simulations aimed at deformable off-road environments.
We assume that robot motion is constrained to the terrain surface, thus defining $\mathbf{s}_\text{start}$ and $\mathbf{s}_\text{goal}$ within a 2.5D spatial space.
$F$ is then given as a unicycle model as follows:
\begin{equation}
\left[\begin{array}{c}
x_{t+1} \\
y_{t+1} \\
\theta_{t+1}
\end{array}\right]=\left[\begin{array}{c}
x_t \\
y_t \\
\theta_t
\end{array}\right]+\Delta t \cdot\left[\begin{array}{c}
{\lambda}_{\text{lin}} \cdot v_t \cdot \cos \left(\theta_t\right) \\
{\lambda}_{\text{lin}} \cdot v_t \cdot \sin \left(\theta_t\right) \\
{\lambda}_{\text{ang}} \cdot \omega_t
\end{array}\right],
\end{equation}
where $\mathbf{s}_t = \left[x_t, y_t, \theta_t\right]^{\top}$, $\mathbf{a}_t = \left[v_t, \omega_t\right]^{\top}$, and ${\boldsymbol{\lambda}} = [{\lambda}_{\text{lin}}, {\lambda}_{\text{ang}}]^{\top}$, simplified as $\lambda = \lambda_{\text{lin}} = \lambda_{\text{ang}}$,
link linear and angular velocities with changes in position and orientation, capturing the dynamics of robot motion.
When formulating the problem, $F$ replaces $\hat{\boldsymbol{\lambda}}$ with $\boldsymbol{\lambda}$ to form eq. (\ref{eq:2e}).
We define $J$ as the function that evaluates states and actions along the trajectory based on their distance to the endpoint for an efficient transition without getting stuck according to the given constraints.
We here exploit $\hat{\Lambda}$ to mark $S_{\text{free}}$ as states where $\hat{\lambda} > \lambda_{\text{stuck}}$, defining $\lambda_{\text{stuck}}$ as the threshold below which a state becomes stuck due to insufficient traversability.
CVaR at level $\alpha=0.9$ quantifies the distribution to assess uncertainty incorporation in motion planning.

As problem solvers, we deploy three planning methods from distinct domains, each designed for off-road navigation.
\subsubsection{Search-based method}

Following \cite{fan2021step}, we implement a hierarchical method divided into global path planning with A* search~\cite{hart1968formal} and local motion planning using the dynamic window approach (DWA)~\cite{fox1997dynamic}.
For each transition, A* finds the shortest path over long horizons in a 2.5D geometric search space, while DWA takes the resulting waypoints as subgoals to plan feasible actions over short horizons.

\subsubsection{Sampling-based method}

We use CL-RRT~\cite{kuwata2009real} for global motion planning that incrementally expands a tree via feedback control simulation, allowing the computation of dynamically feasible trajectories.
Pure-pursuit and PID controllers manage the steering and velocities of the dynamic model, respectively.
Replanning is triggered when substantial deviations occur between the actual and expected states.

\subsubsection{Optimization-based method}

We adopt MPPI~\cite{williams2017information} for navigation without the use of global path planning.
While a reference path remains beneficial for long-horizon decisions, recent studies favor such waypoint-free navigation to take full advantage of planning in control space.
Terminal cost is given in $J$ to ensure long-term stability and goal alignment over an infinite horizon.

To quantify the performance of these algorithms on given problem instances, we use three key metrics.
\begin{itemize}
    \item \textbf{Success rate (Succ.) [\%]} expresses the ratio of instances for which successful trajectories are found.
    \item \textbf{Total time ($T_{\text{total}}$) [sec]} measures navigation efficiency as the total time taken to traverse a trajectory.
    \item \textbf{Average traversability ($\bar{\lambda}$) [\%]} evaluates navigation safety by averaging all the observed traversability coefficients along a trajectory, with higher values indicating a safer traverse.
\end{itemize}

\subsection{Experimental Setups}

We prepare three environmental scenarios: standard (Std), harsh geometry (HG), and harsh shading (HS).
The Std scenario adopts a simple setting mirroring the geometry and appearance trends from ML model training.
The HG scenario introduces unique geometric conditions resulting in low traversability, with crater-like terrain in addition to random elevation changes.
The HS scenario offers unique appearance conditions complicating traversability prediction due to ambiguous visual cues from variable shading.
Each problem instance takes a $32 \times 32\, \text{m}$ map with a $0.5\, \text{m}$ resolution.
Off-road navigation starts at $\mathbf{s}_{\text{start}} = (8\, \text{m}, 8\, \text{m})$ and continues until $\mathbf{s}_{\text{goal}} = (24\, \text{m}, 24\, \text{m})$.

\subsection{Results}

We highlight representative simulation examples in the Std scenario by displaying snapshots of ongoing navigation at $t=25\, [\text{sec}]$ using distinct planning methods, as illustrated in Fig.~\ref{fig:2}.
The nature of these planning algorithms is evident in their trajectories: Fig.~\ref{fig:2}a shows a safe and fairly efficient trajectory under A* guidance; Fig.~\ref{fig:2}b a suboptimal trajectory due to its sampling nature; and Fig.~\ref{fig:2}c an efficient but slightly hazardous trajectory, a consequence of its short-horizon solutions.
We emphasize BenchNav's ability to apply various planning algorithms, focusing on synthetic data generation and probabilistic traversability prediction with built-in ML models, as fundamental steps prior to motion planning. 
This feature allows users to easily simulate off-road navigation, facilitating problem formulation and implementation of planning algorithms while accounting for uncertain traversability.

We also present a quantitative summary of MPPI planning to see how different environmental scenarios impact off-road navigation performance.
Table \ref{tab:quantitative_results} summarizes the results from simulation experiments across 20 problem instances for each of the Std, HG, and HS scenarios, with $T=100\, [\text{sec}]$.
MPPI performs best in the Std scenario, where ML models provide accurate traversability predictions due to training on in-domain data.
The HG and HS scenarios introduce challenges resulting in lower success rates and longer traverse times as ML models struggle with out-of-domain features such as steep inclines and shaded visuals.
In off-road environments, robots often face novel situations that lead to \emph{epistemic uncertainty}, stemming from insufficient training data.
Thus, incorporating uncertainty is critical in assessing the risk inherent in erroneous traversability prediction.
BenchNav allows us to implement various planning algorithms and explore risk inference metrics for resilient off-road navigation.
It equips well-controlled environmental feature generation to replicate challenging scenarios, providing insight into the factors contributing to uncertain traversability predictions.

\subsection{Limitations and Possible Extensions}
\label{sec:limitations_and_possible_extensions}

We clarify BenchNav's limitations as 1) the lack of local observability using onboard sensors and 2) insufficient 3D representations in both perception for handling overhanging objects and motion for representing dynamic robot behaviors.
We will implement these capabilities to simulate navigation algorithms designed for more dynamic, aggressive maneuvers in unknown environments~\cite{han2023model,meng2023terrainnet}.

\begin{table}[t]
    \centering
    \caption{MPPI planning results on different scenarios}
    \begin{tabular}{l|ccc}
        \toprule
        Scenario & Succ. & $T_{\text{total}}$ & $\bar{\lambda}$ \\
        \midrule
        Std & 90 &  52.7 $\pm$ 22.5 & 76.3 $\pm$ 8.8 \\
        HG & 80 &  60.2 $\pm$ 26.1 & 76.5 $\pm$ 4.5 \\
        HS & 85 &  58.6 $\pm$ 25.7 & 74.2 $\pm$ 11.4 \\
        \bottomrule
    \end{tabular}
    \label{tab:quantitative_results}
    \vspace{-5pt}
\end{table}

\section{Conclusion}

We have developed BenchNav, a simulator that tests planning algorithms alongside ML-based traversability modeling for off-road navigation.
Simulation experiments verified that BenchNav consistently executes different off-road navigation algorithms and provides well-controlled datasets replicating difficulties present in traversability prediction.
Future work will address the limitations described in \ref{sec:limitations_and_possible_extensions} to make the simulator more realistic. 
We also aim to extend navigation components by adopting end-to-end learning~\cite{cai2022risk,cai2022probabilistic,cai2023evora,castro2023how} and reinforcement learning techniques~\cite{gan2022energy,triest2023learning,margolis2023learning}.


\bibliographystyle{IEEEtran}
\bibliography{IEEEexample}

\end{document}